\documentclass{bmvc2k}

\title{Exploiting Multiple Priors for Neural 3D Indoor Reconstruction}

\addauthor{Federico Lincetto}{federico.lincetto@phd.unipd.it}{1}
\addauthor{Gianluca Agresti}{Gianluca.Agresti@sony.com}{2}
\addauthor{Mattia Rossi}{Mattia.Rossi@sony.com}{2}
\addauthor{Pietro Zanuttigh}{zanuttigh@dei.unipd.it}{1}

\addinstitution{
 Media Lab,\\
 University of Padova,\\
 Padova, IT
}
\addinstitution{
 Stuttgart Laboratory 1,\\
 Sony Semiconductor Solutions Europe,\\
 Sony Europe B.V.,\\
 Stuttgart, GE
}

\runninghead{Lincetto et al.}{Exploiting Multiple Priors for Neural 3D Indoor Rec.}

\usepackage{amssymb}
\usepackage[acronym]{glossaries}
\usepackage{multirow}
\usepackage{wrapfig}
\newacronym{sdf}{SDF}{Signed Distance Field}
\newacronym{mlp}{MLP}{Multi-Layer Perceptron}
\newacronym{sfm}{SfM}{Structure-from-Motion}
\newacronym{roi}{RoI}{Region of Interest}
\newacronym{pdf}{PDF}{Probability Distribution Function}
\newacronym{mvs}{MVS}{Multi-View Stereo}
\usepackage{comment}

\begin{document}

\maketitle

\begin{abstract}
\noindent Neural implicit modeling permits to achieve impressive 3D reconstruction results on small objects, while it exhibits significant limitations in large indoor scenes. In this work, we propose a novel neural implicit modeling method that leverages multiple regularization strategies to achieve better reconstructions of large indoor environments, while relying only on images. A sparse but accurate depth prior is used to anchor the scene to the initial model. A dense but less accurate depth prior is also introduced, flexible enough to still let the model diverge from it to improve the estimated geometry. Then, a novel self-supervised strategy to regularize the estimated surface normals is presented. 
Finally, a learnable exposure compensation scheme permits to cope with challenging lighting conditions.
Experimental results show that our approach  produces state-of-the-art 3D reconstructions in challenging indoor scenarios.
\end{abstract}

\section{Introduction}
\label{sec:intro}

Recent developments in neural implicit representation strategies permit to build dense geometric models of scenes. These techniques can not only produce continuous representations, but they also exhibit very promising results in contexts very challenging for traditional 3D reconstruction methods, e.g., texture-less areas. 
Among these models, NeRF-based approaches, enhanced with more stringent geometrical constraints, showed impressive results on the 3D reconstruction of object-centric scenes~\cite{DVR,yariv2020multiview,oechsle2021unisurf}. However, the geometry estimation of fine structures is still challenging due to the nature of implicit representation models which tend to ignore high frequency details.
Many NeRF-based solutions rely on geometry cues from depth sensors or external methods to guide the reconstruction~\cite{roessle2022depthpriorsnerf,Azinovic_2022_CVPR}  but, in these cases, the results are strongly dependent on the quality of the geometrical hints. 
Furthermore, these methods struggle on larger and indoor scenes, due to the higher geometry complexity. In such scenarios, acquisitions may capture regions at different scales and focus on many complex structural details. 
Moreover, in real scenarios, brighter or darker regions of the scene can cause the camera exposure to change significantly: this results in the same subject exhibiting very different colors across different images.

In this work, we present a NeRF-based method which learns an implicit representation of indoor environments and produces accurate 3D reconstructions. Our method exploits external depth priors and a novel self-supervised normal regularization. In particular, we introduce a soft depth supervision which suggests to the model the geometry of the scene indirectly, by optimizing the sampling procedure. Moreover, we regularize the surfaces through a self-supervised strategy which exploits spatial information available in the input images that are first pre-processed by a learnable exposure compensation scheme. Reconstruction results outperform competing methods on the 3D reconstruction of indoor scenes.
\begin{figure}[t]
    \centering
    \includegraphics[width=1\textwidth,trim=0 0.3cm 0 0.2cm,clip]{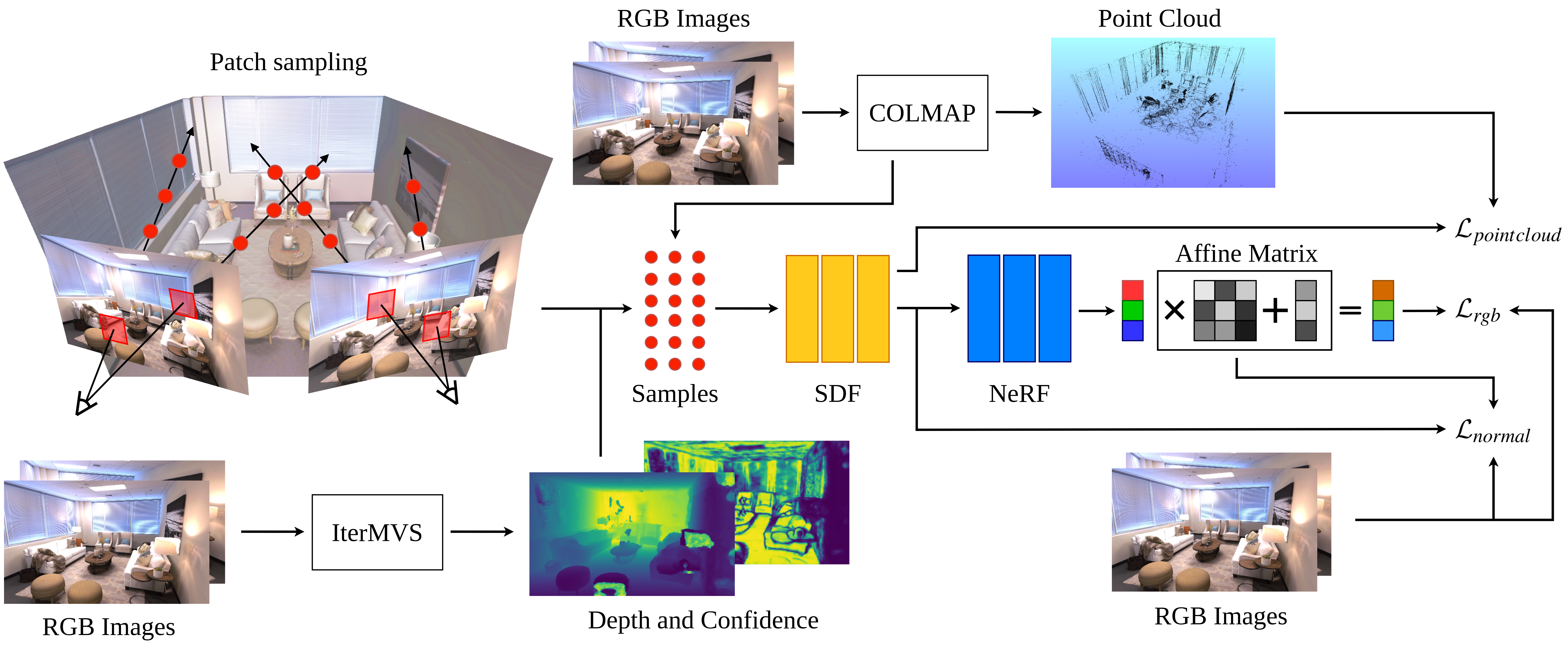}
    \vspace{-0.3cm}
    \caption{Overview of the proposed approach.}
    \label{fig:visual_abstract}
    \vspace{-0.5cm}
\end{figure}

\vspace{-0.2cm}
\section{Related Work}
\label{sec:related}
\vspace{-0.3cm}
In recent years, neural implicit representations~\cite{chen2019learning} have gained popularity and have spread across many different domains~\cite{michalkiewicz2019implicit,dupont2021coin,yuce2022structured,peng2021neural,strumpler2022implicit,saragadam2022miner}. Geometry reconstruction is one of the fields that has benefited the most from these approaches. 
Traditional explicit representation for 3D scenes, like voxel grids~\cite{gadelha20173d,jimenez2016unsupervised},  point clouds~\cite{achlioptas2018learning,camuffo2022recent} or  meshes~\cite{kato2018neural,wang2018pixel2mesh} have some drawbacks, such as the high memory footprint, the fixed resolution and their discrete nature.

Implicit representations aim at solving many of the above mentioned limitations. 
They can represent a continuous volume, so the extraction of explicit models at any resolution becomes possible. Moreover, they drastically reduce the required memory footprint. In the field of 3D representations, the first proposed solutions represented the volume exploiting \glspl*{sdf}~\cite{park2019deepsdf,zhang2021learning} or occupancy grids~\cite{mescheder2019occupancy,oechsle2021unisurf}. \Gls*{sdf} suddenly became the choice for many works aimed at obtaining accurate geometry reconstruction of objects or scenes due to its differentiable nature. Recently, new solutions based on parametric encodings have been published, showing impressive results~\cite{muller2022instant,chen2022tensorf,chen2023factor,wang2022neus2}. The idea is to learn the \gls*{sdf} by substituting the network or enhancing its capability through local parameters stored in multi-resolution data structures. At the price of a higher memory footprint, these approaches can improve the reconstruction quality and reduce the training time.

In the literature there are examples of methods exploiting external geometrical clues~\cite{Yu2022MonoSDF,wang2022neuris,Atzmon_2020_CVPR,roessle2022depthpriorsnerf} to guide the reconstruction. Knowing in advance the depth or the normal of some surface points adds strong constraints that help the model to converge faster and to estimate a better geometry. These priors could come from depth sensors~\cite{zanuttigh2016tof} or they may be estimated through monocular or \gls*{mvs} methods~\cite{wang2022itermvs,sormann2021ib}. Other approaches exploit point clouds from \gls*{sfm} approaches for supervision: the \gls*{sdf} estimation can be guided by these point clouds~\cite{kangle2021dsnerf,Fu2022GeoNeus} or it is possible to initialize a per-point feature vector encoding color and density information~\cite{xu2022point}.

There exist also SDF-based methods explicitly targeting  indoor scenes: specifically, NeuRIS~\cite{wang2022neuris} is a recent method that employs monocular normal maps to guide the reconstruction relying also on a patch-match regularization strategy; MonoSDF~\cite{Yu2022MonoSDF} leverages the information in  monocular depth maps in addition to  normal maps to enforce stronger geometrical constraints. With respect to these methods, our approach exploits external priors by proposing a depth map soft supervision which focuses the sampling on the area closer to the surface without introducing a direct supervision. Furthermore, the proposed normal self regularization acts as smoothing constraint without the need for external normal maps that are often inaccurate in challenging scenarios. Finally, a revised exposure compensation approach improves the results in challenging lighting conditions, typical of indoor scenes.
\vspace{-0.2cm}
\section{Method}
\label{sec:method}
The goal of this work is to reconstruct the geometry of complex indoor environments starting from RGB images only. Previous SDF-based approaches for the task, like NeuS~\cite{wang2021neus} and IDR~\cite{yariv2020multiview}, focus on learning the geometry and the appearance of object-centric scenes. We introduce new supervision and regularization strategies to improve the reconstruction accuracy of larger scenes.
More in detail, we propose two depth-based supervision strategies exploiting sparse and dense depth priors, respectively (Sec.~\ref{sec:depth_supervision}).
Furthermore, we introduce a strategy to handle variable exposure settings (Sec.~\ref{sec:exposure_compensation}) and a  self-supervised method to regularize the surface normals  (Sec.~\ref{sec:normal_regularization}).

\vspace{-0.2cm}
\subsection{Neural Implicit Surface Rendering}
\label{sec:neural_implicit_representations}

\paragraph{Scene Representation}
The geometry is represented using the Signed Distance Function (\gls*{sdf}). This is a function $f: \mathbb{R}^3 \to \mathbb{R}$ that assigns to every point in the volume the distance between this point and the closest surface in the scene. Consequently, the scene surface $\mathcal{S}$ is the set of 3D coordinates with \gls*{sdf} value equal to 0: formally it is $\mathcal{S} = \left\{x \in \mathbb{R}^3 | f\left( x \right) = 0 \right\} $.

Moreover, the radiance field is represented as a function $c: \mathbb{R}^3 \times \mathbb{S}^2 \to \mathbb{R}^3$ that assigns to every 3D point a color depending on the camera viewing direction, with $\mathbb{S}^2$ defined as {$\mathbb{S}^2=\left\{x \in \mathbb{R}^3:\lVert x\rVert=1\right\}$}. The functions $f$ and $c$ are modeled by a couple of \glspl*{mlp}.

\paragraph{Volume Rendering} Since the aim of this work is accurate geometry reconstruction, we follow the common volume rendering solution but substituting the density term with a more geometrically constrained one, called opacity $\rho$. Following the parameterization proposed in NeuS~\cite{wang2021neus}, this term is a function of the Sigmoid computed at the \gls*{sdf} values. This formulation is unbiased and occlusion aware, making the method better suited for 3D Reconstruction.
Once the opacity is computed, the rendering procedure is the same as in the original NeRF~\cite{mildenhall2020nerf}. All the colors and opacities along each ray are summed up to produce the final color according to the following relation:
\vspace{-0.1cm}
\begin{equation}
    C(\mathbf{o}, \mathbf{v}) = \int_{0}^{+\infty} w(t)c(\mathbf{p}(t), \mathbf{v})  \,dt
\end{equation}
where $\mathbf{o}$ is the position of the camera, $\mathbf{v}$ the viewing direction, $\mathbf{p} = \mathbf{o} + t\mathbf{v}$ the 3D point and $c$ and $w$ the color and the weight at each specific coordinate, respectively. The weight term is defined as $w(t) = T(\rho(t))\rho(t)$, with $T$ representing the accumulated transmittance. 
We refer the reader to NeuS~\cite{wang2021neus} for further details.

\vspace{-0.2cm}
\subsection{Variable Exposure Compensation}
\label{sec:exposure_compensation}
While  object-centric scenes considered in previous approaches exhibit uniform illumination, larger indoor scenes often have many light variations and  are often acquired with variable camera exposure settings. In order to compensate for the differences in colors coming from the non-constant exposure and white balancing, inspired by the method proposed in \cite{rematas2022urf}, we introduce an exposure compensation strategy based on an affine model. 
In particular, an affine transformation is learned for each training image and applied to the RGB color produced by the rendering network before comparing it with the ground truth.

Differently from~\cite{Azinovic_2022_CVPR, martinbrualla2020nerfw}, we do not employ ad-hoc compensation networks to avoid an over-parameterization of the model~\cite{rematas2022urf} and we directly optimize the 12 parameters of the affine transformation. Therefore, the rendered color $C_k(p)$  for pixel $p$ in image $k$ is:
\begin{equation}
   C_k(p) =  R_k \hat{C}_k(p) + \mathbf{t}_k \quad\text{with}\quad A_k= [R_k|\mathbf{t}_k]
\end{equation} 
where $\hat{C}_k(p)$ is the output of the rendering network.
The parameters are initialized with $R_k=I$ and $\mathbf{t}_k=\mathbf{0}$.
Such an initialization scheme is an initial strong prior but it is reasonable since the real colors are not too far from the acquired ones. Furthermore, in order to avoid the convergence to unexpected color shifts, we introduce an anchor point. In particular, the matrix, corresponding to the reference image chosen among the ones with the most uniform color histogram, is fixed to the initialization values.
In this way, the affine matrices corresponding to all the other images are forced to produce a color appearance resembling that of the reference image, thus aligning the exposure settings of the whole scene.

\vspace{-0.2cm}
\subsection{Depth Supervision}
\label{sec:depth_supervision}

Since our model relies exclusively on RGB images, we propose to exploit geometry information retrieved from the same RGB pictures used as input. Such supervision is beneficial since it supports the \gls*{sdf} network training by adding reliable geometrical constraints that improve the results and a guidance for the sampling stage to speed-up the training. Querying the \gls*{sdf} model permits producing information about the depth value for each pixel, that can be compared with depth data retrieved from RGB images through other strategies.

We implement two synergic supervision strategies: a point cloud supervision, exploiting sparse depth data, and a depth map soft supervision, employing dense depth data.

\paragraph{Sparse Pointcloud Supervision}
One approach, similar to DS-NeRF~\cite{kangle2021dsnerf}, exploits the sparse point cloud produced by the \gls*{sfm} algorithm during the camera pose estimation. In our setting we use COLMAP~\cite{schoenberger2016mvs,schoenberger2016sfm}. Generally, these 3D points (denoted keypoints) are sparse but reliable, since they are the result of a robust feature extraction and triangulation procedure.
For this purpose, it is possible to consider the rays passing through the point cloud keypoints seen by the current camera \cite{kangle2021dsnerf}. At this point, the weights $w$ of the points along the ray are used to estimate a depth value as follows:
\begin{equation}
    \label{eq_depth}
    \hat{D}(r) = \int_{t_n(r)}^{t_f(r)} w(t)tdt
\end{equation}
where $r$ is the considered ray, $t_n(r)$ and $t_f(r)$ the near and far \gls*{roi} bounds along the considered ray, respectively, and $t$ the distance of that sample from the camera position. This permits to define a loss $\mathcal{L}_{pc}$ that compares the depth $\hat{D}$ rendered by the model against the depth $D$ predicted by COLMAP.

It is important to notice that the produced point cloud is sparse therefore, when projected on the current camera image plane, most of pixels do not observe a keypoint. Considering that each pixel of an image may be chosen to cast a ray, this kind of supervision could be rarely applicable. For this reason, we choose to implement the supervision in parallel with the standard NeRF pipeline: we randomly cast a batch of rays for the main pipeline and a different batch of rays through the pixels observing a keypoint. These two batches are processed independently. Finally, we choose to consider only keypoints seen by at least a minimum number of different cameras, for better robustness.

\paragraph{Depth Map Soft Supervision}
The point cloud supervision is effective in supporting the geometry estimation, but it is based on sparse data. Usually, less than 1000 valid keypoints are available for each camera, hence the majority of the rays is not supervised by the point cloud. 
For this reason, the next step aims at implementing a supervision that is available at any pixel. We adopt an approach exploiting depth and confidence maps generated by IterMVS~\cite{wang2022itermvs}, a deep-learning algorithm for \gls*{mvs} depth map estimation.
The main idea behind this approach is to exploit this depth information to focus the sampling only in the region close to the surface. To achieve this goal, a novel soft supervision strategy is presented. It is based on a \gls*{pdf}, built on the \gls*{sdf} weights $w$, which expresses the probability to encounter a surface at each step along a ray. 

In the first phase, the NeuS original coarse-to-fine sampling is performed, in order to produce a \gls*{pdf} $h(x)$ which locates the estimated surfaces along the ray. 
This PDF is the result of the interpolation of the weights $w$ along the ray. At this point, according to depth information acquired from depth maps, a Gaussian \gls*{pdf} $n(x)=\mathcal{N}(d, \sigma)$ is generated. 
This is centered at the depth values $d$ and  has standard deviation $\sigma$ proportional to the confidence $\mu$ of depth estimation. This \gls*{pdf} is combined with the \gls*{pdf} estimated by the \gls*{sdf} network:
\begin{equation}
    g(x) = h(x)  n(x) = h(x)  \mathcal{N}(d,\sigma) \;\;\;\; \mathrm{with} \;\;\;\; \sigma = 0.5  (1 - \mu).
\end{equation}
Note that, if the two PDFs are independent, the product represents the joint probability. In our case, $h(x)$ is the probability of finding a surface along a ray, while $n(x)$ can be seen as the probability that the estimated depth is coherent with  the depth map. Moreover, they come from different computations, so they can be considered independent.

\begin{figure}[]
    \centering
    \begin{minipage}{0.60\textwidth}
    \resizebox{0.93\textwidth}{!}{
        \includegraphics[trim=0 0.1cm 0 0,clip]{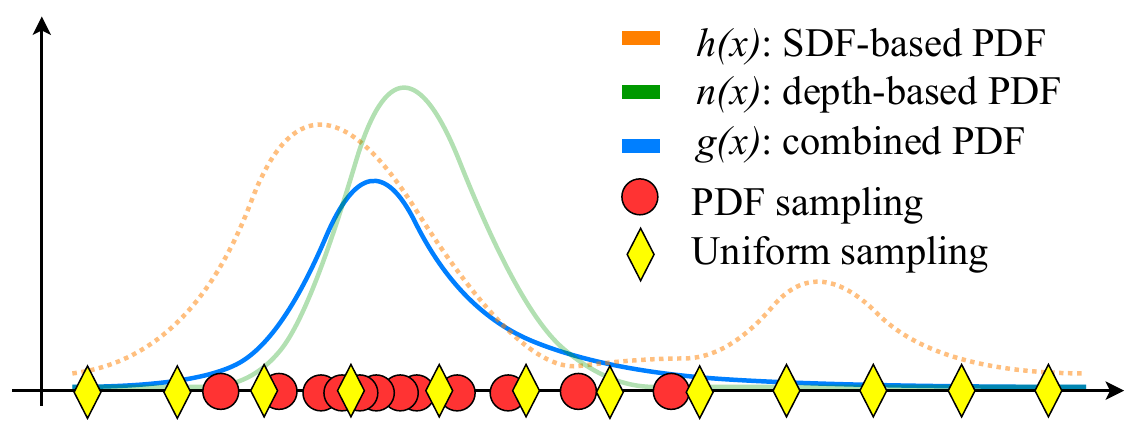}}
        \caption{Depth map guided sampling strategy.}
        \label{fig:depth_supervision}
    \end{minipage}
    \begin{minipage}{0.38\textwidth}
    \resizebox{0.93\textwidth}{!}{
        \includegraphics[trim=0 0.5cm 0 0,clip]{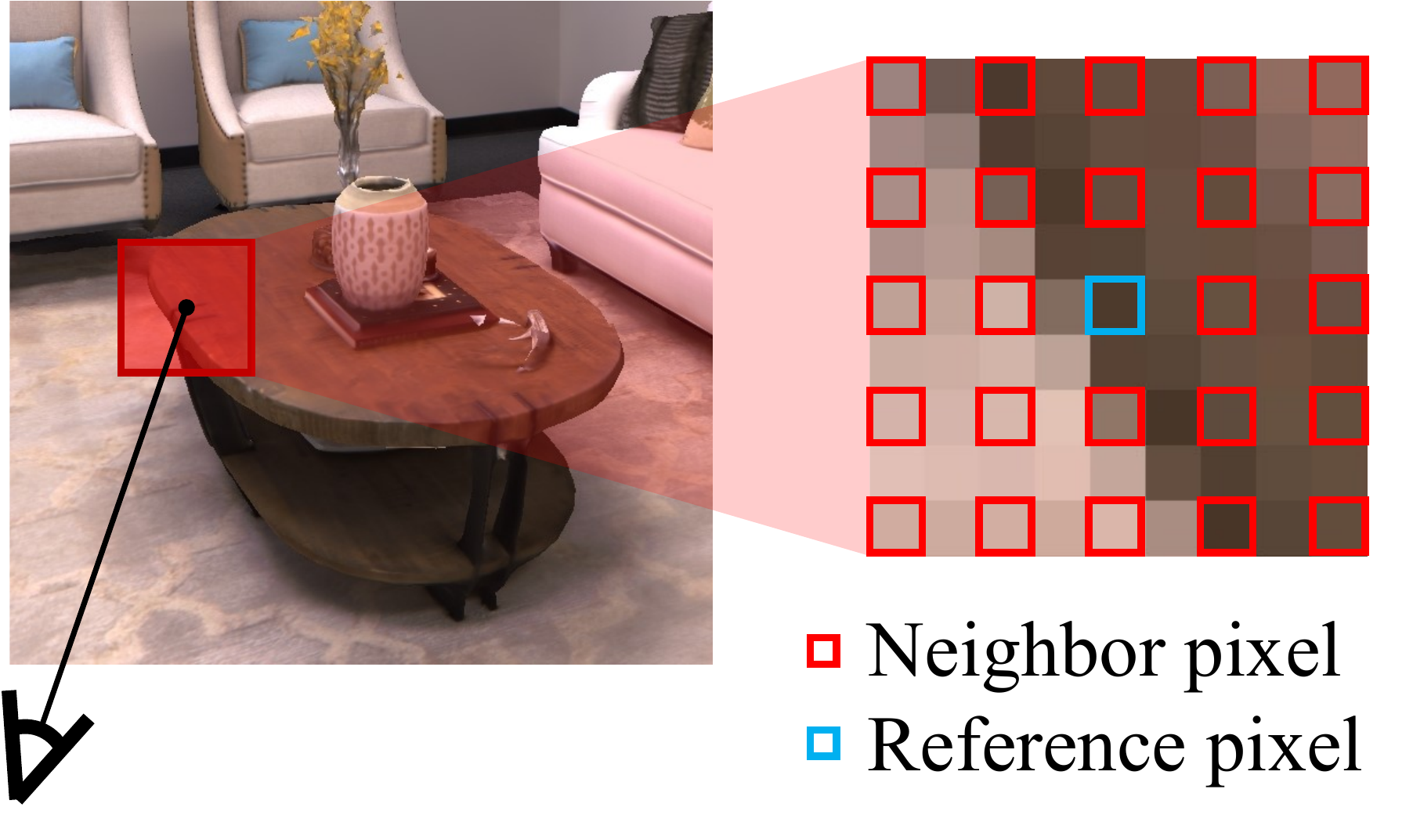}}
        \caption{Normal regularization.}
        \label{fig:normal_regularization}
    \end{minipage}
    \vspace{-15pt}
\end{figure}

Finally, all previous points are discarded and the new \gls*{pdf} $n(x)$ is used to re-sample less but more relevant new points, which will be used for the rendering pipeline (see Fig.~\ref{fig:depth_supervision}). More in detail, half of these points (marked with ``{\color{black}$\circ$}'' in the example of Fig.~\ref{fig:depth_supervision}) are sampled according to $g(x)$, the second half is sampled linearly along the ray (``{\color{black}$\diamond$}'' in the figure). 
This permits an accurate sampling close to the surface for increased precision, and a coarse sampling on all the ray to handle critical cases where the current surface is wrong.

As a final consideration, in principle, the depth maps could be used to define an explicit loss, as done for the point cloud supervision. We evaluated this option: however since the depth information from these maps is generally less accurate, employing an explicit loss would lead to instability during the training, with different depth supervisions pulling towards different directions. For this reason, we choose to keep the more reliable depth supervision on point clouds and employ depth maps only to guide the sampling.

\vspace{-0.2cm}
\subsection{Normal Self Regularization}
\label{sec:normal_regularization}
Even if both depth supervisions are effective in adding a reliable guidance to the geometry estimated by the \gls*{sdf}, it may happen that texture-less planar surfaces are not reconstructed as smooth as expected. This can be explained considering that COLMAP and IterMVS struggle to extract and triangulate reliable features on texture-less areas, leading to a lack of supervision in those regions.

To address this problem, we present a novel normal self regularization strategy. The idea is to guide the geometry estimation without using external normal priors, generally costly to compute and not always reliable. 
In our implementation, the model  exploits both the information available in RGB images and in depth data. In the standard setting, at every iteration, sparse random pixels are sampled: however, this does not permit to exploit the spatial consistency due to the sparsity of the rays. Our solution consists in considering patches of pixels instead of single sparse pixels, and to cast a batch of rays for every patch, thus exploiting spatial information. The rationale is that close pixels with a similar color should represent surface points of a common plane~\cite{rossi2020joint}. Otherwise, when the color is different, we cannot make any assumption on the geometry.
This idea is implemented by computing for every pixel representing a surface point the surface normal, the surface depth and a  weight. The depth is computed as described in Eq.~\ref{eq_depth}, while the surface normal can be estimated as the gradient of the \gls*{sdf} in that specific point:
\vspace{-0.2cm}
\begin{equation}
    \hat{N}(r) = \int_{t_n(r)}^{t_f(r)} w(t)\nabla \mathbf{p}(t)dt
\end{equation}
where $r$ is the considered ray, $t_n$ and $t_f$ the near and far \gls*{roi} bounds and $t$ the distance of that sample from the camera position.
Then, as shown in Fig.~\ref{fig:normal_regularization}, we consider the central pixel of every patch as the reference and compute the weights $w_b$ quantifying the likelihood that one surface point belongs to the same plane of the central one, similarly to a bilateral filter:
\begin{equation}
    w_b(r) = \mathcal{N}(I(i) - I(j), \sigma_c)  \mathcal{N}(\hat{P}(i) - \hat{P}(j), \sigma_d)
\end{equation}
where $\mathcal{N}$ is the normal distribution while $I$ is the pixel color corrected with the respective exposure compensation matrixes of Sec.~\ref{sec:exposure_compensation} and converted to the CIE Lab color space to better capture the color similarity. Finally, $P$ are the 3D coordinates of the surface point and $i$ and $j$ the neighbor and reference pixels, respectively. The bilateral weights are computed considering the difference in color and in the actual location in the scene of two surface points. Therefore, the model can regularize wide surfaces while preserving edges and high frequency details. At this point, it is possible to force the surface normals to be similar to the central point surface normal according to the respective bilateral weight. 
The spatial term is needed to enforce that the regularization is applied to points on the same surface, thus avoiding to regularize points with the same color but far from each other.

\vspace{-0.2cm}
\subsection{Optimization}
\label{sec:optimization}
The loss function used to optimize our model is composed by several terms, as follows:
\vspace{-0.1cm}
\begin{equation}
    \mathcal{L} = \mathcal{L}_{rgb} + \lambda_1\mathcal{L}_{eikonal} + \lambda_2\mathcal{L}_{pointcloud} + \lambda_3\mathcal{L}_{normal}
\end{equation}

\vspace{-0.4cm}

\paragraph{Photometric Loss}
The first contribution is given by the photometric loss, computed comparing the rendered RGB pixels against the image ones. It is defined as follows:
\vspace{-0.1cm}
\begin{equation}
    \mathcal{L}_{rgb} = \frac{1}{m} \sum_{r \in \mathcal{R}} \left\lvert (R_k\hat{C}_r + \mathbf{t}_k) - C_r \right\rvert
\end{equation}
where $\mathcal{R}$ is the set of rays, $m\!\!=\!\!|\mathcal{R}|\!$  the number of rays,  $\hat{C}$ the estimated colors with the exposure compensation terms for current image $k$ (see Sec.~\ref{sec:exposure_compensation}) and $C$ the ground truth ones.

\vspace{-0.3cm}
\paragraph{Eikonal Loss}
The Eikonal loss enforces that the gradient of the \gls*{sdf} has unitary norm:
\vspace{-0.1cm}
\begin{equation}
    \mathcal{L}_{eikonal} = \frac{1}{nm} \sum_{ \mathbf{p} \in \mathcal{X}} \left( \left\lVert \nabla f( \mathbf{p}) \right\rVert _2 - 1 \right)^{2}
\end{equation}
where $n$ is the sampling bin size and $\mathcal{X}$ is the set of all sampled points along the rays.

\paragraph{Point Cloud Loss} The depth loss is the contribution accounting for the point cloud supervision. It enforces that the estimated depth rendered by the density values is close to the depth predicted by COLMAP keypoints. It is defined as follows:
\begin{equation}
    \mathcal{L}_{pointcloud} = \frac{1}{m}\sum_{i \in \mathcal{K}} w_i \left\lvert \hat{D}_{r_i} - D_i\right\rvert ^2
    \label{eq:pointcloud_loss}
\end{equation}
where $\mathcal{K}$ is the set of COLMAP keypoints seen by the current camera, $r_i$ is the ray passing through the point $i$ and $D_i$ the depth of the $i$-th keypoint.

\vspace{-0.2cm}
\paragraph{Normal Loss} The normal loss aims at smoothing the surfaces as result of the normal self regularization strategy. It forces the normals of neighbor pixels to be similar to the normal of the central reference pixel in case they lay on the same surface. It is defined as follows:
\vspace{-0.1cm}
\begin{equation}
    \mathcal{L}_{normal} = \frac{1}{m} \sum_{\mathcal{B} \in \mathcal{P}}  \sum_{r \in \mathcal{B}} \lVert \hat{N}_r - \hat{N}_j \rVert * w_{b}(r)
\end{equation}
where $m$ is number of rays cast per image,  $\mathcal{P}$ is the set of pacthes and $\mathcal{B}$ is the set of rays in a single patch.

\vspace{-0.2cm}
\section{Experiments}
\label{sec:experiments}

In this section, we present the results achieved by our method, denoted ``MP-SDF'', and compare them with the state-of-the-art in the field.
We employ the Replica~\cite{replica19arxiv} dataset: this contains real world indoor scenes acquired by RGB-D cameras and provides ground truth meshes for quantitative evaluations. 
Furthermore, in order to better evaluate the impact of the exposure compensation on scenes with a more challenging illumination, we test our method also on the \texttt{Meetingroom} scene from the Tanks and Temples~\cite{Knapitsch2017} dataset. This scene is significantly larger than the Replica ones, thus making it an interesting test case.
The evaluation is performed using the Chamfer-\textit{L}$_1$ distance and the F-score as quantitative metrics, following the experimental protocol of MonoSDF~\cite{Yu2022MonoSDF}.

\subsection{Implementation Details}
\label{sec:implementation_details}

\begin{figure}[]
    \centering
    \includegraphics[width=\textwidth]{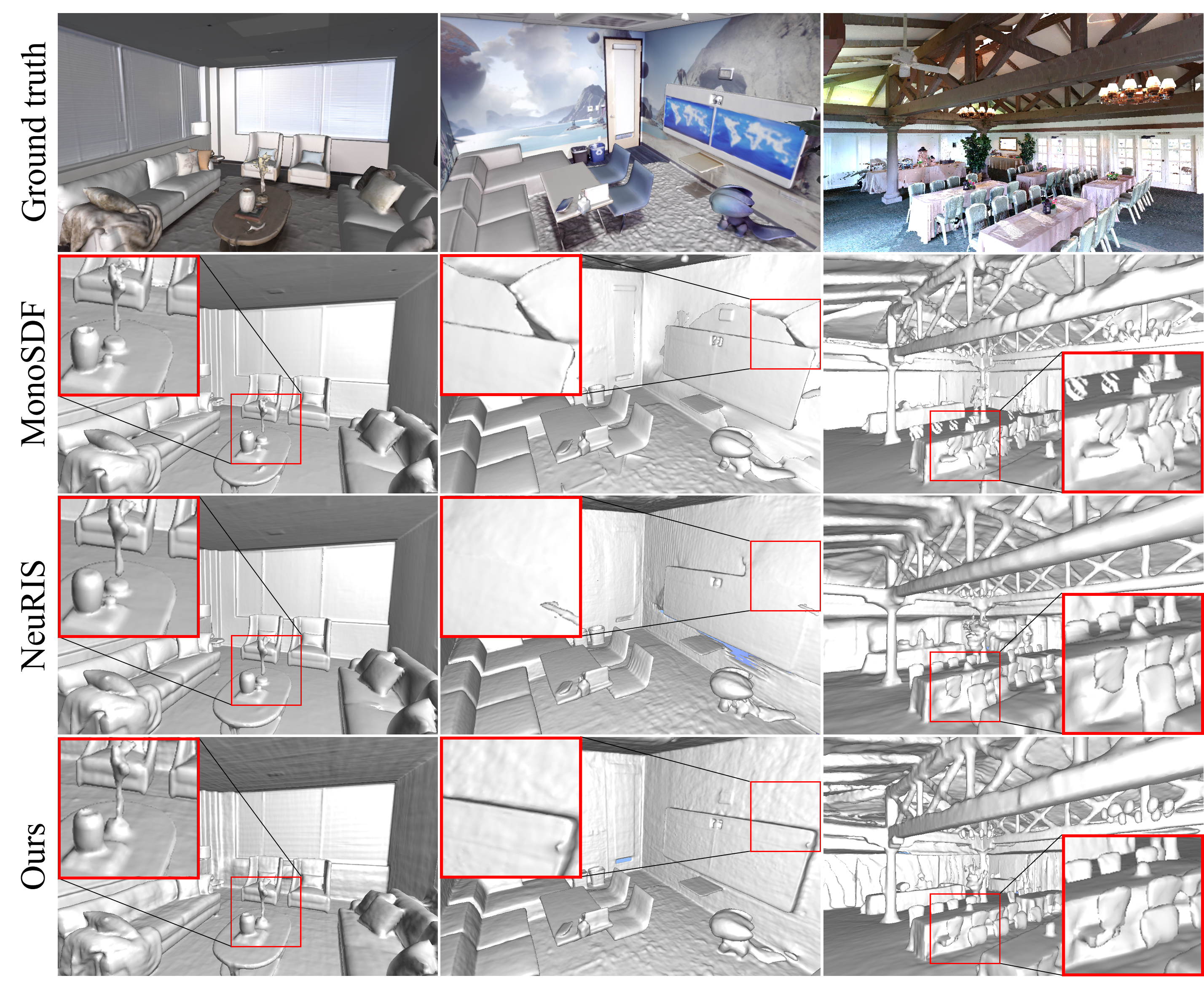}
    \caption{Qualitative comparison between the reconstructed meshes of Replica \texttt{scan1} and \texttt{scan4} and of Tanks and Temples' \texttt{Meetingroom} by MonoSDF, NeuRIS and our method. The ground truth is colorful to contextualize the scene, while reconstruction are colorless to better appreciate geometry details. The red boxes underline the more accurate reconstruction achieved by our approach.}
    \vspace{-5pt}
    \label{fig:results}
\end{figure}

\begin{table}[]
\resizebox{\textwidth}{!}{
    \begin{tabular}{c | l c c c c c c c c | c || c c c c c c c c | c}
        \hline
            && \multicolumn{9}{c||}{GT Poses} & \multicolumn{9}{c}{COLMAP Poses}\\
        \cline{2-20}
            & scan & 1 & 2 & 3 & 4 & 5 & 6 & 7 & 8 & Mean & 1 & 2 & 3 & 4 & 5 & 6 & 7 & 8 & Mean\\
    
         \hline
            \multirow{4}{*}{\rotatebox[origin=c]{90}{Chamfer $\downarrow$}} &
            MonoSDF Grids & \textbf{2.22} & 2.27 & \textbf{1.63} & 3.53 & 4.36 & \textbf{2.00} & \textbf{2.16} & 2.21 & 2.55 & \textbf{2.23} & 2.64 & \textbf{1.62} & 3.15 & 4.44 & \textbf{2.03} & \textbf{2.19} & 2.22 & 2.56 \\  
            & MonoSDF MLP & 2.26 & 2.03 & 1.81 & 3.97 & 5.00 & 2.10 & 2.27 & 2.32 & 2.72 & 2.35 & 1.88 & 1.80 & 3.99 & 5.10 & 2.25 & 2.26 & 2.37 & 2.75\\
            & NeuRIS & 2.48 & 2.46 & 1.74 & 2.39 & 4.05 & 2.11 & 3.51 & \textbf{1.98} & 2.59 & 2.83 & 2.15 & 1.73 & 3.23 & 5.48 & 2.05 & 3.07 & \textbf{1.95} & 2.81\\
         \cline{2-20}
            & \textbf{MP-SDF (Ours)} & 2.60 & \textbf{1.75} & 1.74 & \textbf{1.58} & \textbf{2.14} & 2.32 & 2.80 & 2.30 & \textbf{2.15} & 3.00 & \textbf{1.64} & 1.80 & \textbf{2.12} & \textbf{2.60} & 2.19 & 2.94 & 2.20 & \textbf{2.31}\\
         \hline
            \multirow{4}{*}{\rotatebox[origin=c]{90}{F-score $\uparrow$}} &
            MonoSDF Grids & 94.95 & 92.67 & \textbf{99.05} & 86.52 & 77.16 & 94.45 & 93.74 & 94.05 & 91.57 & \textbf{94.87} & 92.95 & \textbf{99.16} & 86.49 & 76.60 & \textbf{95.02} & 95.54 & 93.75 & 91.80\\  
            & MonoSDF MLP & \textbf{95.36} & 96.20 & 98.72 & 84.94 & 72.29 & 95.03 & \textbf{95.75} & 93.75 & 91.51 & 93.34 & 96.98 & 97.99 & 83.82 & 71.54 & 92.07 & \textbf{95.83} & 92.82 & 90.55\\
            & NeuRIS & 92.83 & 95.10 & 97.75 & 91.48 & 75.23 & 94.76 & 90.88 & \textbf{97.69} & 91.97 & 88.56 & 95.70 & 97.99 & 86.22 & 66.42 & 94.87 & 88.95 & \textbf{97.93} & 89.58\\
         \cline{2-20}
            & \textbf{MP-SDF (Ours)} & 91.71 & \textbf{96.35} & 97.26 & \textbf{97.47} & \textbf{92.34} & \textbf{95.17} & 90.49 & 91.78 & \textbf{94.41} & 89.70 & \textbf{98.10} & 96.31 & \textbf{92.69} & \textbf{88.08} & 94.53 & 90.17 & 94.23 & \textbf{93.44}\\
         \hline
    \end{tabular}
}
\vspace{2pt}
\caption{Quantitative comparison of the accuracy on Replica dataset.}
\label{tab:results}
\end{table}

\begin{figure}[]
    \centering
    \includegraphics[width=\textwidth]{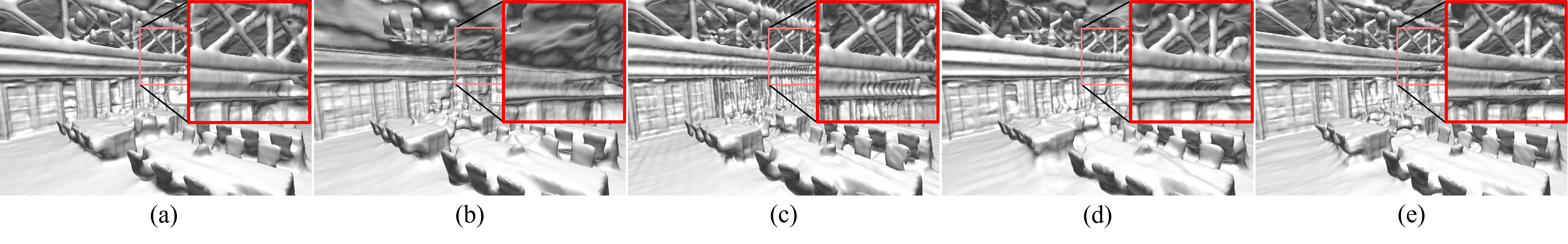}
    \vspace{-0.8cm}
    \caption{Ablation study results on \texttt{Meetingroom}. (a): full model, (b): w\textbackslash o point cloud, (c): w\textbackslash o depth map, (d): w\textbackslash o exposure correction, (e): w\textbackslash o normal regularization.}
    \vspace{-15pt}
    \label{fig:ablation}
\end{figure}

We implement our method on top of NeuS~\cite{wang2021neus}. The \gls*{sdf} function $f$ and the color $c$ are estimated by a \gls*{mlp} with 12 and 8 hidden layers, respectively. The \gls*{sdf} \gls*{mlp} is initialized to predict a sphere~\cite{Atzmon_2020_CVPR}. 
The network input dimensionality is augmented by the positional encoding~\cite{mildenhall2020nerf} with 10 and 4 frequencies for the 3D coordinates and the viewing direction, respectively. The model is trained for 300k iterations on an Nvidia A6000  using the Adam optimizer. 
The learning rate warmup proposed in NeuS~\cite{wang2021neus} is maintained. The warmup training stage consists of 60k iterations while the refinement stage covers the remaining iterations. In both stages, $1008$ rays are cast and $256$ points are sampled with the NeuS coarse-to-fine procedure, while 128 points are re-sampled by the soft depth supervision. The number of rays cast through the point cloud depends on the valid keypoints, while the maximum number of these rays is set to 128.
The patch size and dilation rate for  normal regularization are set to 3$\times$3 and $2$, respectively. The bilateral weights are computed by setting $\sigma_c=3 $ and $\sigma_d=0.03$. 
In the sparse point cloud supervision, keypoints seen by less than 5 cameras are discarded. The  loss weights are empirically set to $\lambda_1=\lambda_3=0.1$ while  $\lambda_2$ starts at $0.5$ and decreases exponentially during the training, thus the point cloud can help at the beginning to initialize the model correctly, while in further stages the model can differentiate more from the initial prior.

\vspace{-0.2cm}
\subsection{3D Reconstruction Experiments}
\label{sec:3d_experiments}
We compare our results against MonoSDF~\cite{Yu2022MonoSDF}, the current state-of-the-art NeRF-based method for indoor geometry reconstruction, and against NeuRIS~\cite{wang2022neuris}, an improvement of NeuS  exploiting normal maps for indoor reconstruction. MonoSDF relies on monocular depth and normal priors, while NeuRIS exploits monocular normal maps only, with a patch-match strategy to discard inaccurate normal priors.
To provide geometric priors to these methods we employ Omnidata~\cite{eftekhar2021omnidata,kar20223d}, as suggested by MonoSDF authors.

We test the methods starting from both images with ground truth and estimated poses: the results on the Replica dataset are shown in Tab.~\ref{tab:results}. 
On average, our model  achieves better results in both cases and according to both metrics.
With ground truth poses, the average F-score  is $94.41$ against $91.97$ of the best competitor. Using  COLMAP poses, results are similar: $93.44$ against $91.80$. The Chamfer distances also confirm these results. 
More in detail, for some scenes MP-SDF produces better reconstructions, while for the others the results are close to competitors.
Looking at Fig.~\ref{fig:results}, it is clear how MonoSDF and NeuRIS produce good results in some environments but fail in some others, while our approach exhibits more stable results across all scenes. As an example, on Replica \texttt{scan4}, the wall behind the whiteboard presents some artefacts when reconstructed by concurrent methods, while MP-SDF reconstructs it correctly. 
This happens because the competitors rely heavily on input data from external monocular estimators that cannot offer strong reliability: these data may be inconsistent between subsequent frames and they may fail in case of prospectively misleading backgrounds, such as the painted wall in the second row. 
In our case, the depth maps are estimated by IterMVS, which is a \gls*{mvs} method, thus ensuring consistency over different images. See the supplementary material for more details. 

Moreover, notice how the Omnidata model used by competitor methods to extract monocular clues was pre-trained on a poll of datasets, including Replica itself. 
To ensure fairness, the tests should be done on scenes not included in the Omnidata training.
Therefore, we perform an additional test on the \texttt{Meetingroom} scene of the Tanks and Temples dataset.
The results achieved by our model outperform the reconstructions of the other two methods, achieving a F-score of $52.53$ and a Chamfer distance of $10.02$. MonoSDF achieves F-score and Chamfer distance of $45.03$ and $11.22$, respectively, when trained with the multi-resolution grids and $40.45$ and $13.77$ with the MLP training.  NeuRIS reaches a F-score of $37.13$ and a Chamfer distance of $12.32$. 
The fact that our method can outperform competitors even with a more ``fair'' training procedure is another proof of its value.

\vspace{-0.2cm}
\subsection{Ablation Study}

We propose an ablation study on the employed supervision strategies to investigate the relevance of every contribution on the reconstruction quality. For this purpose, we  show the results achieved by our ablated model on  the \texttt{scan1} and  \texttt{Meetingroom} scenes.  
\begin{wraptable}{r}{0.45\textwidth}
    \resizebox{0.45\textwidth}{!}{
    \begin{tabular}{c | l c c }
        \hline
            && \texttt{Scan1} & \texttt{Meetingroom}\\
    
        \hline
            \multirow{5}{*}{\rotatebox[origin=c]{90}{F-score $\uparrow$}} &
            MP-SDF (full model) & 94.40 & 52.53\\
            & w\textbackslash o point cloud & 93.14 & 39.25\\
            & w\textbackslash o depth map & 90.18 & 47.90\\
            & w\textbackslash o exposure correction & 94.20 & 39.36\\
            & w\textbackslash o normal regularization & 92.55 & 46.63\\
        \hline
    \end{tabular}}
    \caption{Ablation studies.}
    \label{tab:ablation}
\end{wraptable}
The results are presented in Tab. \ref{tab:ablation} and in Fig. \ref{fig:ablation}.

All the supervision strategies contribute to improve the reconstruction quality. The two depth supervisions are effective in improving the geometry accuracy: without the point cloud supervision the reconstructed geometry is severely affected by a lack of guidance, leading to estimate wrong structures. On the other hand, the soft depth map supervision helps in removing periodic artifacts on surfaces, introducing a smoothing effect. The normal self-regularization permits to obtain a general smoothing of the surfaces, removing bumps but preserving the edge sharpness. The tests on the model with the exposure correction disabled show that it generally improves the reconstruction. This is more relevant in the \texttt{Meetingroom} scene, since it was acquired with variable exposure settings. On Replica \texttt{scan1}, acquired with fixed exposition and white balance, the exposure compensation effects are less relevant.

\vspace{-0.4cm}
\section{Limitations}
\label{sec:limitations}
\vspace{-0.2cm}
State-of-the-art competitors like MonoSDF and NeuRIS compute losses from a direct comparison with the depth and/or normal maps generated with a monocular estimator. Thus, if these prior maps are very accurate, they produce good reconstructions. Differently, in MP-SDF the direct depth loss of the point cloud supervision affects only some pixels due to the point cloud sparsity, resulting in a softer guidance. 
Furthermore, MP-SDF depth soft supervision helps to focus the sampling but it does not enforce a direct constraint on the geometry.
Similarly, MP-SDF normal regularization smooths artifacts out, but it does not directly force normal orientation.
For these reasons, if the monocular priors are very accurate, our model could be outperformed by competitors. 
Generally, the monocular estimators produce very smooth maps, leading to reconstruct also smoother mesh with respect to our method which does not rely on monocular priors. Anyway, when monocular maps contain errors their reconstruction is strongly affected, as shown in Fig. \ref{fig:results}. 

\vspace{-0.4cm}
\section{Conclusion}
\label{sec:conclusion}
\vspace{-0.2cm}
In this work we introduce a novel approach targeting the reconstruction of complex indoor scenes. The joint usage of multiple priors, including soft depth supervision and advanced normal regularization strategies permits to achieve state-of-the-art results in 3D reconstruction.
Future research will be devoted to the employment of multi-resolution schemes and to the reduction of the computational requirements.
\vspace{-0.4cm}
\paragraph{Acknowledgment} This collaborative work was funded by Sony Europe B.V.. Special thanks go to Oliver Erdler, Yalcin Incesu and Piergiorgio Sartor for their support.


\clearpage
\appendix
\section*{\huge Supplementary Material}
\noindent In this document we provide some additional details in order to better evaluate the performances of  MP-SDF. 
Firstly, we present in detail the employed comparison metrics (Section~\ref{sec:metrics}).
Then, a further analysis is shown to justify the use of a \gls*{mvs} method rather than a monocular one for depth priors (Section~\ref{sec:prior_evaluation}). Moreover, we discuss the exposure compensation effectiveness showing the results achieved on scenes acquired with variable or fixed exposure settings and then reconstructed with or without exposure compensation enabled (Section~\ref{sec:exposure_correction}).
Finally, we present more qualitative results showing the reconstructed scenes (Section~\ref{sec:qualitative}).
\vspace{-0.2cm}
\section{Quantitative Metrics}
\label{sec:metrics}

In this section, we recall the definition of the quantitative metrics employed in this work. In particular, we used the F-score and the Chamfer-\textit{L}$_1$ distance. These two metrics compare the ground truth point cloud against the predicted one, extracted considering the set of vertexes of the respective mesh. The predicted mesh is produced by MP-SDF querying the SDF network on a three dimensional grid defined at a chosen resolution, then running the marching cubes algorithm~\cite{lorensen1987marching}. In our experiments, aligned to what done in MonoSDF, the meshes are extracted from a grid at a resolution of $512\times512\times512$. 

\vspace{-0.2cm}
\paragraph{F-score} To define  F-score, it is helpful to introduce  precision $p$ and  recall $r$ first. Precision is defined as follows:
\vspace{-0.2cm}
\begin{equation}
    p = \frac{1}{\left|\mathcal{P}\right|} \sum_{\mathbf{p} \in \mathcal{P}} \mathbf{1}_T\left( \, \text{min}_{\mathbf{p^*} \in \mathcal{P}^*} \lVert \mathbf{p} - \mathbf{p^*} \rVert \, \right) \quad \text{with} \quad \mathbf{1}_T(x):=
    \begin{cases}
        1 \,\, \text{if} \,\, x\,<\,T\\
        0 \,\, \text{if} \,\, x\,\ge\,T
    \end{cases}
    \vspace{-0.2cm}
\end{equation}
where $\mathcal{P}$ is the predicted point cloud, $\mathcal{P}^*$ the ground truth one and $T=0.05$. Recall is defined as follows:
\begin{equation}
    r = \frac{1}{\left|\mathcal{P}^*\right|} \sum_{\mathbf{p^*} \in \mathcal{P}^*} \mathbf{1}_T\left( \, \text{min}_{\mathbf{p} \in \mathcal{P}} \lVert \mathbf{p} - \mathbf{p^*} \rVert \, \right).
\end{equation}
At this point, it is possible to define the F-score as follows:
\vspace{-0.2cm}
\begin{equation}
    \text{F-score} = \frac{2 \cdot p \cdot r}{p+r}.
\end{equation}
The F-score values presented in the main article are multiplied by a factor 100, to express them as percentages.

\vspace{-0.2cm}
\paragraph{Chamfer-\textit{L}$_1$ distance} To define the Chamfer-\textit{L}$_1$ distance it is convenient to introduce the accuracy $a$ and the completeness $c$ first. The accuracy is defined as follows:
\vspace{-0.2cm}
\begin{equation}
    a = \frac{1}{\left|\mathcal{P}\right|} \sum_{\mathbf{p} \in \mathcal{P}} \left( \, \text{min}_{\mathbf{p^*} \in \mathcal{P}^*} \lVert \mathbf{p} - \mathbf{p^*} \rVert \, \right)
\end{equation}
where $\mathcal{P}$ is the predicted point cloud and $\mathcal{P}^*$ the ground truth one. The completeness is defined as follows:
\begin{equation}
    c = \frac{1}{\left|\mathcal{P}^*\right|} \sum_{\mathbf{p^*} \in \mathcal{P}^*} \left( \, \text{min}_{\mathbf{p} \in \mathcal{P}} \lVert \mathbf{p} - \mathbf{p^*} \rVert \, \right).
\end{equation}
At this point, it is possible to define the Chamfer-\textit{L}$_1$ distance as:
\begin{equation}
    \text{Chamfer-\textit{L}$_1$} = \frac{a+c}{2}.
\end{equation}
The F-score assigns a fixed weight to points with a distance larger than the selected threshold, independently on the actual error. On the contrary, the Chamfer-\textit{L}$_1$ distance is computed considering every distance value, which makes it more sensible to outliers.
\section{Depth Prior Evaluation}
\label{sec:prior_evaluation}

In this section we propose a further investigation on the use of external priors. MonoSDF~\cite{Yu2022MonoSDF} and NeuRIS~\cite{wang2022neuris} exploit monocular depth and normal maps produced by Omnidata~\cite{eftekhar2021omnidata} to supervise the training. Instead, MP-SDF exploits multi-view stereo depth maps obtained from IterMVS~\cite{wang2022itermvs}. 

\begin{figure}
    \centering
    \includegraphics[width=\textwidth]{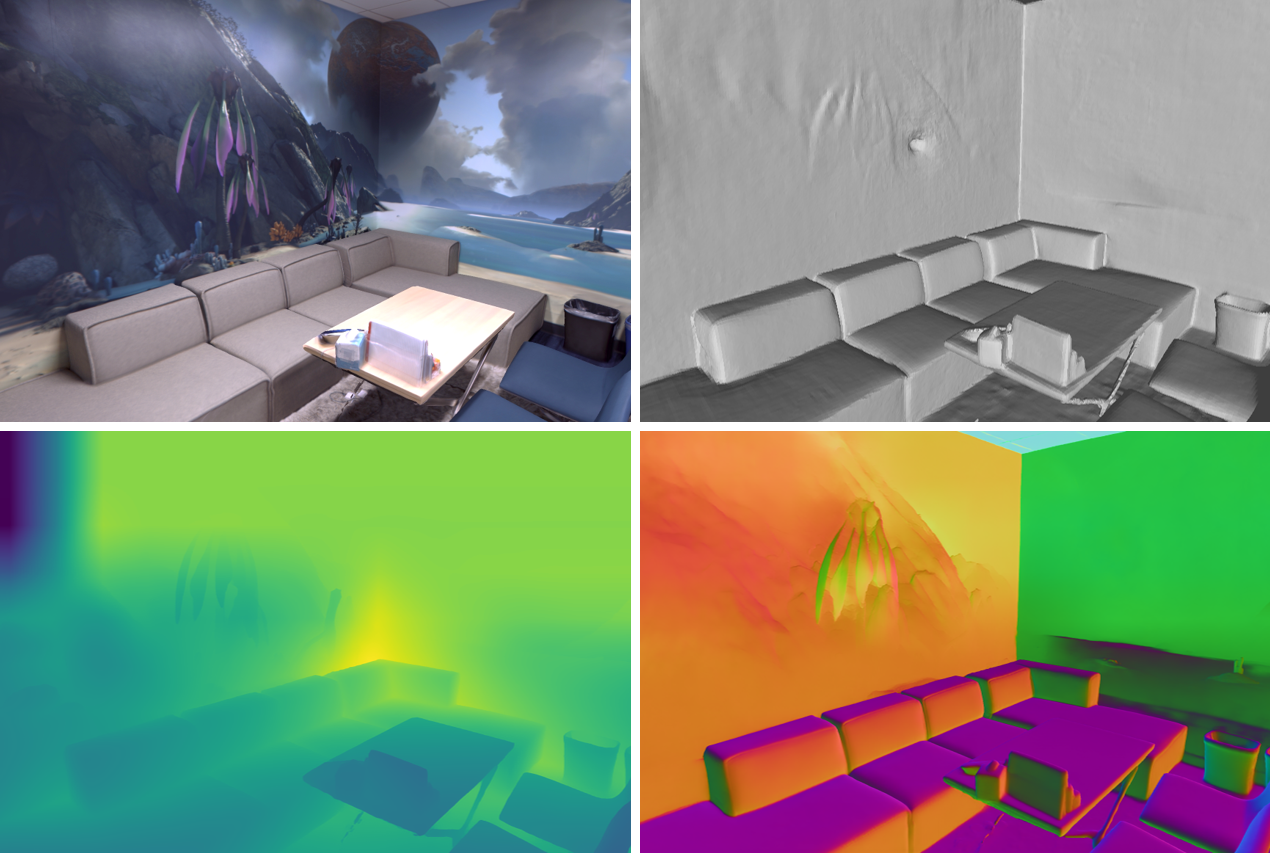}
    \caption{Replica \texttt{scan4}. From left to right and from top to bottom: RGB ground truth, MonoSDF reconstructed mesh, Omnidata depth map, Omnidata normal map.}
    \label{fig:scan4_omnidata}
\end{figure}

\begin{figure}
    \centering
    \includegraphics[width=\textwidth]{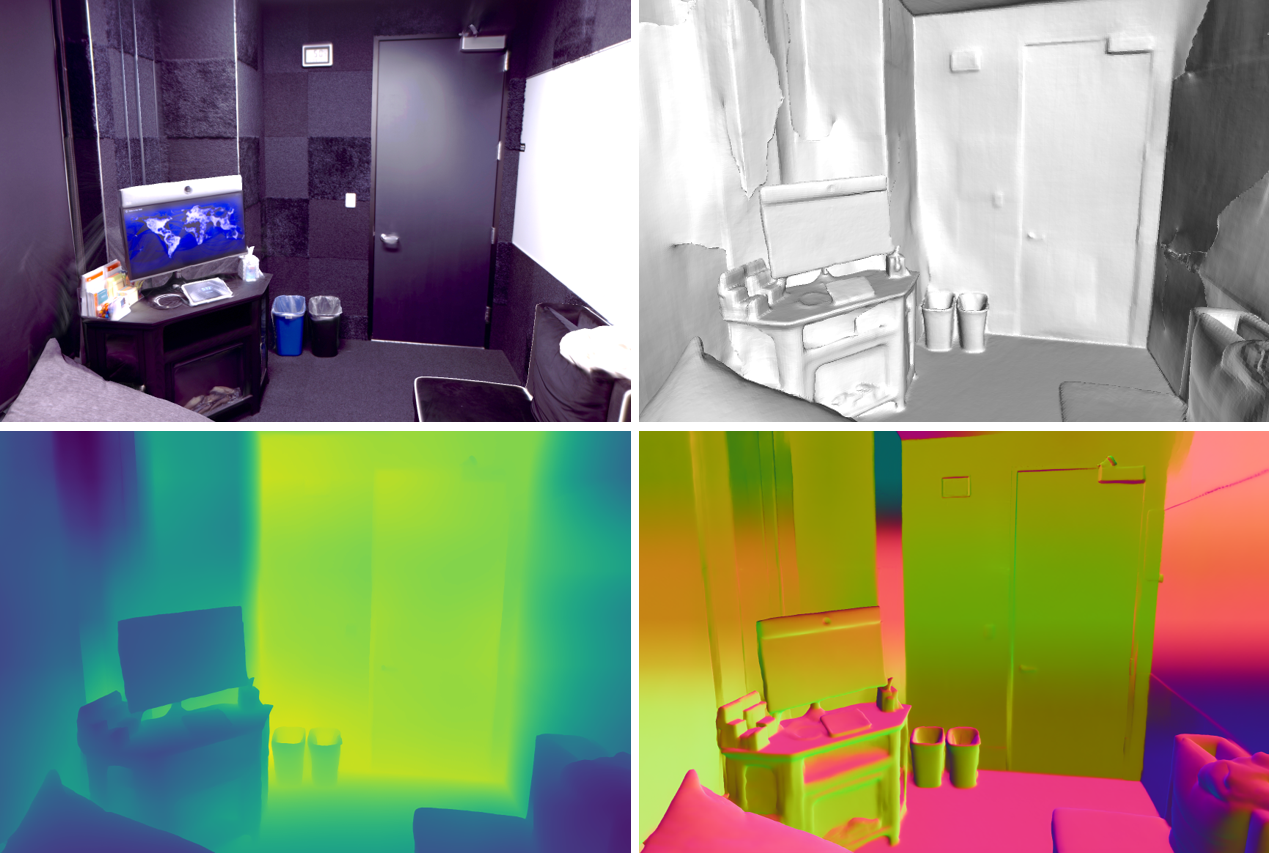}
    \caption{Replica \texttt{scan5}. From left to right and from top to bottom: RGB ground truth, MonoSDF reconstructed mesh, Omnidata depth map, Omnidata normal map.}
    \label{fig:scan5_omnidata}
\end{figure}

We argue that geometric monocular cues can lead to wrong reconstructions: monocular approaches often produce predictions at a wrong scale and not consistent over subsequent frames. Moreover, in presence of flat but perspective coherent backgrounds, they may fail considering the 2D surface as part of the 3D volume. In the case of Replica \texttt{scan4}, presented in Fig.~\ref{fig:scan4_omnidata}, the office wall is painted with a natural background. The monocular depth and normal estimations predict a depth and a normal maps that include geometric details of the background, even if the real surface is flat. Therefore, the reconstructed mesh is biased by these priors and exhibits artifacts on the wall. Considering Replica \texttt{scan5} in Fig.~\ref{fig:scan5_omnidata}, it is clear how Omnidata predicts wrong normals for the office walls. This strongly affects the reconstruction producing wall deformations.

\begin{figure}
    \centering
    \includegraphics[width=\textwidth]{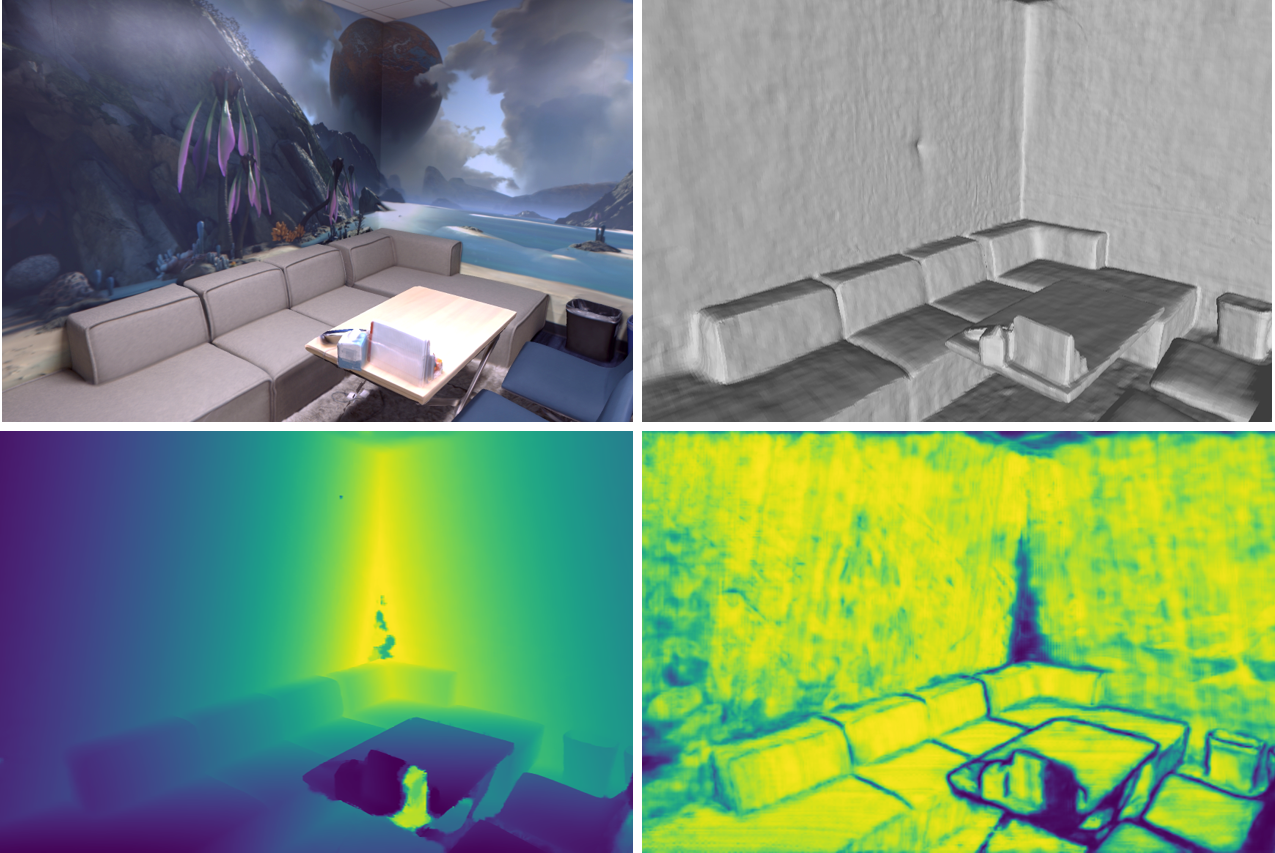}
    \caption{Replica \texttt{scan4}. From left to right and from top to bottom: RGB ground truth, MP-SDF (Ours) reconstructed mesh, IterMVS depth map, IterMVS confidence map.}
    \label{fig:scan4_itermvs}
\end{figure}

\begin{figure}
    \centering
    \includegraphics[width=\textwidth]{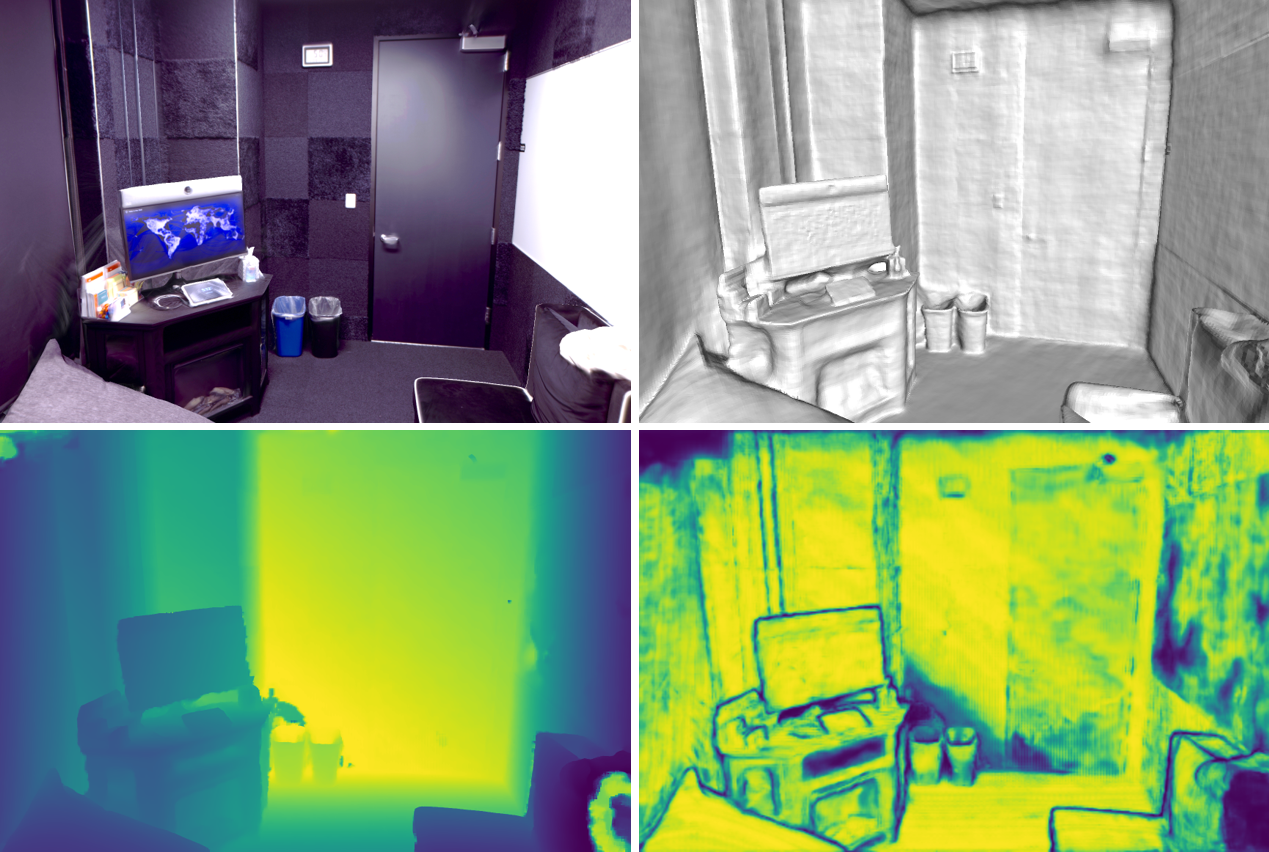}
    \caption{Replica \texttt{scan5}. From left to right and from top to bottom: RGB ground truth, MP-SDF (Ours) reconstructed mesh, IterMVS depth map, IterMVS confidence map.}
    \label{fig:scan5_itermvs}
\end{figure}

Considering the same scenes reconstructed by MP-SDF supervised by IterMVS, it is possible to observe in Fig.~\ref{fig:scan4_itermvs} and Fig.~\ref{fig:scan5_itermvs} that they do not exhibit the artifacts mentioned above. Indeed, the \gls*{mvs} depth maps are more reliable and consistent with the real geometry of the scene. In particular, in \texttt{scan4} the painting does not affect the depth estimation, which is consistent with the planar wall. Finally, in both the considered scenes, the IterMVS confidence maps show high reliability, thus enforcing the quality of \gls*{mvs} predictions.

\section{Exposure compensation effectiveness}
\label{sec:exposure_correction}

\begin{figure}
    \centering
    \includegraphics[width=\textwidth]{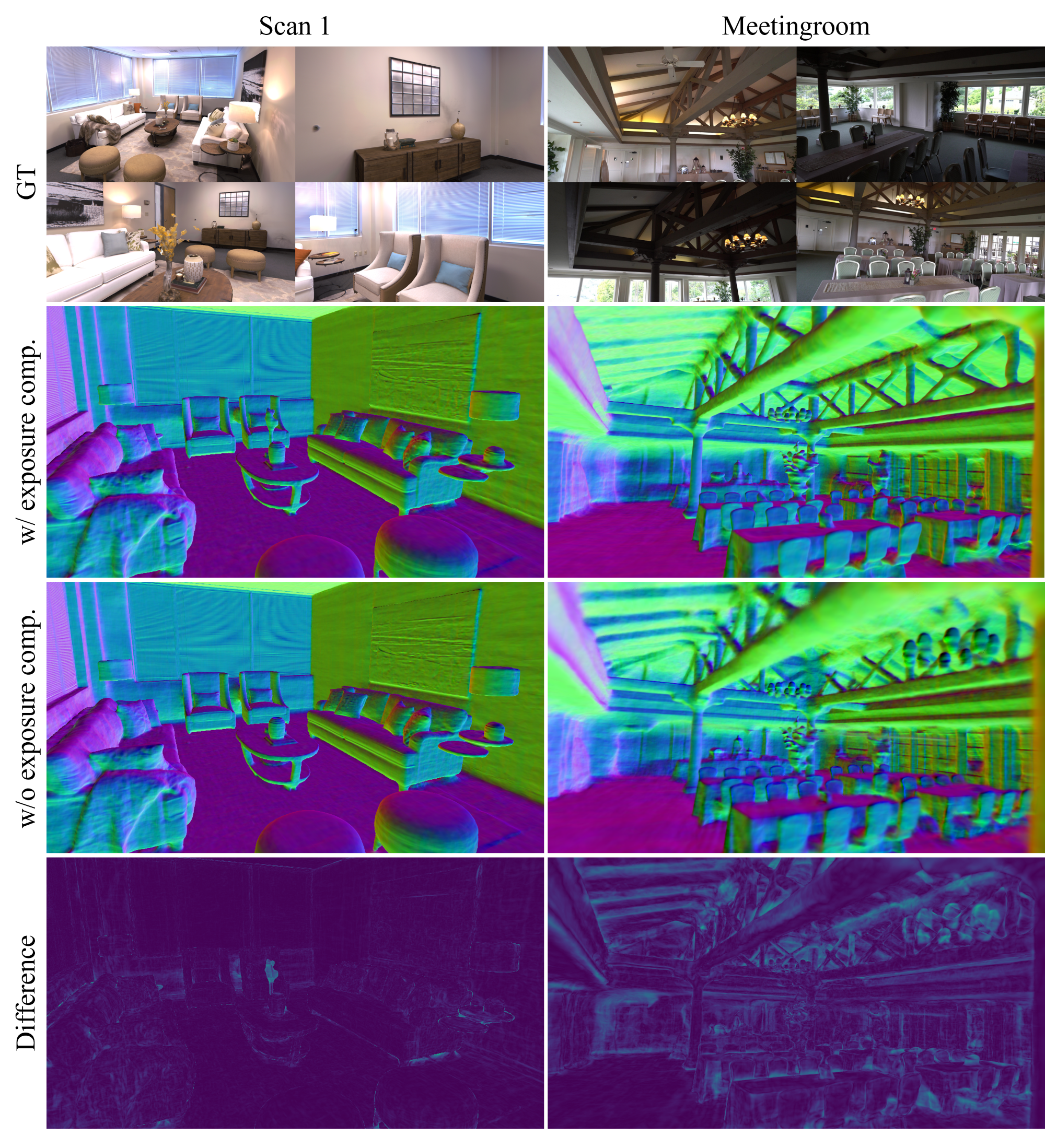}
    \caption{Exposure compensation effectiveness.}
    \label{fig:exposure_correction}
\end{figure}

In Fig. \ref{fig:exposure_correction} we show the effect produced by the exposure compensation. In the first row, it is possible to see some of the input RGB images of the two datasets: the Replica \texttt{Scan 1} is acquired with fixed exposure while the Tanks and Temples \texttt{Meetingroom} is captured with variable exposure. In the second and third rows, the normal maps of the reconstructions obtained with and without the exposure compensation are shown, respectively, while in the fourth row the difference map between normals is presented. As expected, its contribution on Replica is small, since input images were rendered with fixed exposure. Instead, it is very relevant on the Tanks and Temples scene, as it was acquired with variable exposure settings. The exposure compensation helps in reconstructing more accurately the structures on the ceiling and the challenging table and chair area in the middle of the room. As confirmed in Tab. 2 of the paper, the results achieved by the model without the exposure compensation are severely affected on the Tanks and Temples \texttt{Meetingroom}, while on Replica \texttt{Scan 1} the ablation has almost no effect.
\section{Qualitative results}
\label{sec:qualitative}

\begin{figure}
    \centering
    \includegraphics[width=\textwidth]{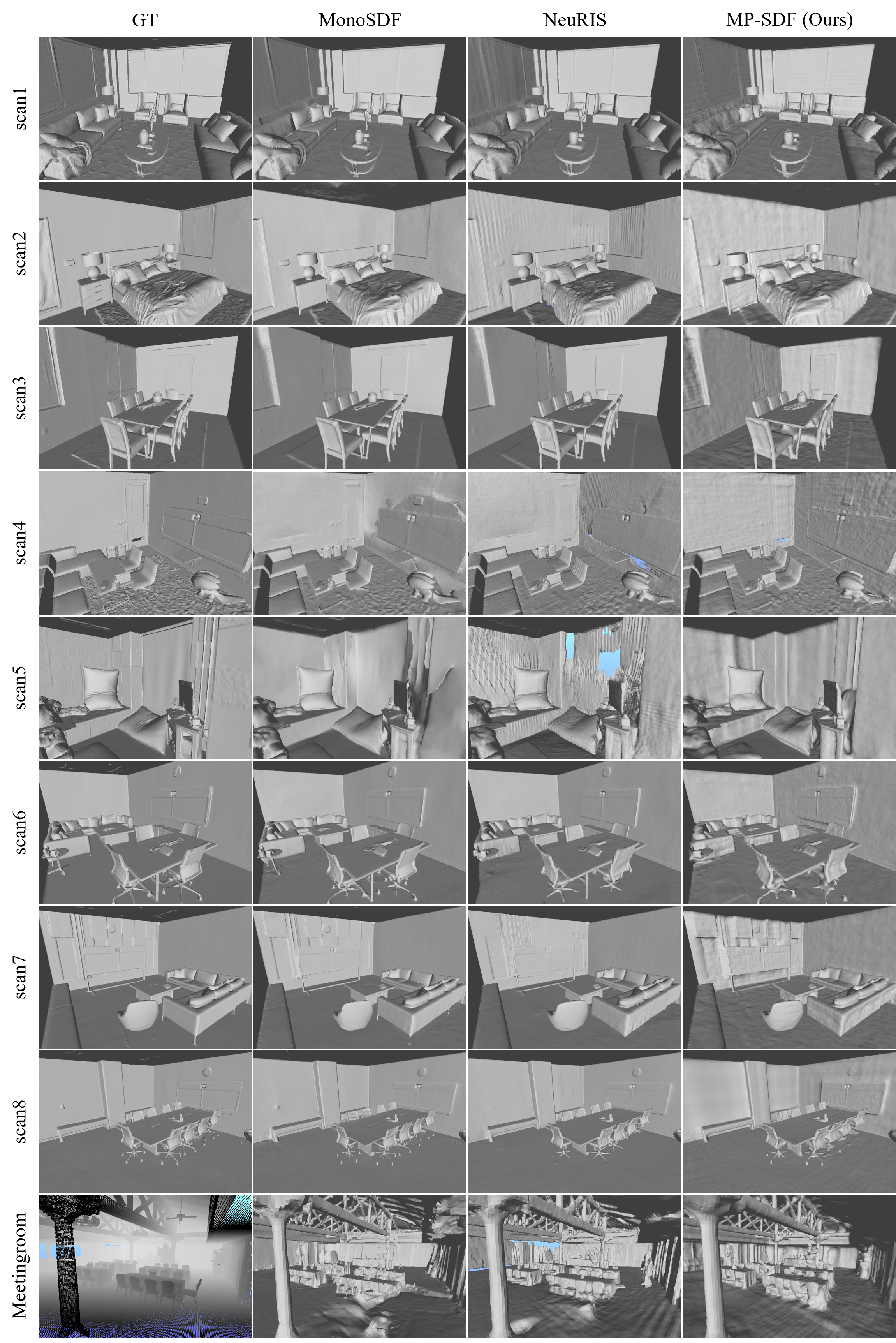}
    \vspace{-0.6cm}
    \caption{Qualitative evaluation on Replica and Tanks and Temples \texttt{Meetingroom}.}
    \label{fig:qualitative}
\end{figure}

Fig.~\ref{fig:qualitative} presents some qualitative comparisons between MonoSDF~\cite{Yu2022MonoSDF}, NeuRIS~\cite{wang2022neuris} and our approach (MP-SDF) on the Replica dataset~\cite{replica19arxiv} and on the \texttt{Meetingroom} scene from Tanks and Temples~\cite{Knapitsch2017}.
As discussed in the main article, on Replica our method outperforms the current state-of-the-art methods for indoor reconstruction, namely MonoSDF and NeuRIS. Differently from them, the visual quality of the reconstruction provided by MP-SDF is consistent across the different scenes in the figure, as also suggested by the quantitative results presented in the article. Finally, MP-SDF clearly outperforms the other methods also on Tanks and Temples \texttt{Meetingroom} in terms of visual quality, also in this case accordingly to the quantitative evaluation in the main article. 

\clearpage

\clearpage

\end{document}